\documentclass[]{antgroup}
\PassOptionsToPackage{numbers, compress}{natbib}
\usepackage{antgroup}

\usepackage{graphicx}
\usepackage{tikz}
\usepackage{todonotes}
\usepackage{multirow}
\usepackage{amsmath}
\usepackage{cleveref}
\usepackage{subcaption}
\usepackage{multicol}
\usepackage{amssymb}
\usepackage{array}
\usepackage{bm}
\usepackage{enumitem}
\usepackage{algorithm}
\usepackage{algpseudocode}
\usepackage{wrapfig}
\usepackage[font=small]{caption} 
\usepackage{tabularx}
\usepackage{bbm}
\usepackage{makecell}
\usepackage{multirow} 
\usepackage{colortbl}
\usepackage{xcolor}

\definecolor{skyblue}{RGB}{0,0,255}   
\definecolor{orange}{RGB}{255,165,0} 
\definecolor{green}{RGB}{0,205,0}

\newcommand{\sys}{dInfer}
\newcommand{\fastdllm}{Fast-dLLM}
\newcommand{\lladamoe}{LLaDA-MoE}

\usepackage[normalem]{ulem}
\useunder{\uline}{\ul}{}

\usepackage{amsmath,amsfonts,bm}

\def\eqref#1{equation~\ref{#1}}

\def\1{\bm{1}}

\DeclareMathAlphabet{\mathsfit}{\encodingdefault}{\sfdefault}{m}{sl}
\SetMathAlphabet{\mathsfit}{bold}{\encodingdefault}{\sfdefault}{bx}{n}

\newcommand*\justify{%
  \fontdimen2\font=0.4em
  \fontdimen3\font=0.2em
  \fontdimen4\font=0.1em
  \fontdimen7\font=0.1em
  \hyphenchar\font=`\-
}

\renewcommand{\texttt}[1]{%
  \begingroup
  \ttfamily
  \begingroup\lccode`~=`/\lowercase{\endgroup\def~}{/\discretionary{}{}{}}%
  \begingroup\lccode`~=`[\lowercase{\endgroup\def~}{[\discretionary{}{}{}}%
  \begingroup\lccode`~=`.\lowercase{\endgroup\def~}{.\discretionary{}{}{}}%
  \catcode`/=\active\catcode`[=\active\catcode`.=\active
  \justify\scantokens{#1\noexpand}%
  \endgroup
}

\usepackage{amsmath}
\usepackage{amssymb}
\usepackage{amsfonts}                               
\usepackage{amsthm}
\usepackage[mathcal]{eucal}
\usepackage{mathrsfs}
\usepackage{bm}                                     
\usepackage{blkarray}                               
\usepackage{nicefrac}                               

\usepackage{wrapfig}
\usepackage{graphicx}                               
\usepackage{caption}
\usepackage{cleveref}

\usepackage{tikz}                                          
\usepackage{circuitikz}
\usetikzlibrary{patterns,snakes}
\usetikzlibrary{positioning,calc,fit,decorations.pathmorphing,shapes.geometric, shapes.gates.logic.US, calc}
\usetikzlibrary{arrows,arrows.meta,decorations.markings,shapes,shapes.arrows}
\usetikzlibrary{decorations,decorations.pathreplacing}
\usetikzlibrary{backgrounds}
\usepackage{filecontents}                           
\usepackage{pgfplots}
\usepackage{pgfplotstable}
\usepgfplotslibrary{groupplots}
\usepackage{scalefnt}
\pgfplotsset{compat=newest}

\usepackage{xcolor}
\definecolor{firstcolor}{HTML}{C3423F}
\definecolor{secondcolor}{HTML}{2A4B8C}

\title{\sys: An Efficient Inference Framework \\ for Diffusion Language Models}

\author{Yuxin Ma$^{1, *}$, Lun Du$^{1, *}$, Lanning Wei$^{1, *}$, Kun Chen$^{1, *}$, Qian Xu$^{1, 4, *}$, Kangyu Wang$^{1, 6, *}$, Guofeng Feng$^{1, 5, *}$, Guoshan Lu$^{1, *}$, Lin Liu$^{1}$, Xiaojing Qi$^{1}$, Xinyuan Zhang$^{1}$, Zhen Tao$^{1}$, Haibo Feng$^{1}$, Zhiyun Jiang$^{1, 3}$, Ying Xu$^{1, 5}$, Zenan Huang$^{1}$, Yihong Zhuang$^{1}$, Haokai Xu$^{1, 2}$, Jiaqi Hu$^{1, 2}$, Zhenzhong Lan$^{1, 3, \dag}$, Junbo Zhao$^{1, 2, \dag}$, Jianguo Li$^{1, \dag}$, Da Zheng$^{1, \dag}$}

\footnotetext{$^\dag$Corresponding Authors. $^*$Core Contributors.}

\affiliation{$^1$Ant Group, $^2$Zhejiang University, $^3$Westlake University, $^4$Renmin University of China, $^5$University of Chinese Academy of Sciences, $^6$Shanghai Jiao Tong University}

\begin{document}

\maketitle
\begin{abstract}

Diffusion-based large language models (dLLMs) have emerged as a promising alternative to autoregressive (AR) LLMs, leveraging denoising-based generation to enable inherent parallelism. Even more and more open-sourced dLLM models emerge, yet their widespread adoption remains constrained by  the lack of a standardized and efficient inference framework. We present \sys, an efficient and extensible framework for dLLM inference. \sys~decomposes the inference pipeline into four modular components--model, diffusion iteration manager, decoding strategy, and KV-cache manager--and integrates novel algorithms for each component alongside system-level optimizations.
Through this combination of algorithmic innovations and system enhancements, \sys~achieves substantial efficiency gains without compromising output quality on \lladamoe. At batch size 1, it surpasses 1,100 tokens per second on HumanEval and averages over 800 tokens per second across six benchmarks on $8\times$ H800 GPUs. Compared to prior systems, \sys~delivers a $10\times$ speedup over Fast-dLLM while maintaining similar model performance.
Even compared to the AR model (with a comparable number of activation parameters and performance) QWen2.5-3B, which is highly optimized with the latest vLLM inference engine, dInfer still delivers a $2$–$3\times$ speedup. The implementation of \sys~is open-sourced at \url{https://github.com/inclusionAI/dInfer}.
\end{abstract}



\section{Introduction}

Over the past year, diffusion-based large language models (dLLMs) have gained increasing attention in both academia and industry. Unlike conventional autoregressive (AR) models that generate tokens sequentially, dLLMs refine entire sequences in parallel through iterative denoising. This intrinsic parallelism opens new opportunities for faster decoding and enables better utilization of GPU hardware. Combined with rapid algorithmic progress, these properties make dLLMs a compelling alternative to AR LLMs. Recent work has shown that models such as LLaDA(-MoE) \citep{nie2025llada,zhu2025lladamoe} can reach performance levels comparable to strong AR baselines, including Llama \citep{grattafiori2024llama3} and Qwen \citep{yang2024qwen2_5}.

Despite these advantages, the practical deployment of dLLMs still faces three critical bottlenecks. First, dLLMs are substantially more computationally expensive than AR models due to iterative denoising steps, making efficiency improvements at both the algorithm and system level essential. Second, while dLLMs have inherent parallelism, scaling parallel decoding remains challenging—larger parallel spans often degrade output quality. Third, the field lacks a unified inference framework and standardized evaluation protocol. This gap hinders consistent benchmarking and often leads to incomparable acceleration claims, such as reporting tokens per second (TPS) under varying batch sizes or hardware.

To address these challenges, we propose \sys, an efficient and extensible inference framework for dLLMs. \sys~modularizes inference into four components--model, diffusion iteration manager, decoding strategy, and KV-cache management--and provides well-designed APIs for flexible combinations of algorithms in each component. It supports multiple dLLM variants, including LLaDA~\citep{nie2025llada}, \lladamoe~\citep{zhu2025lladamoe}, and \lladamoe-TD (Section \ref{sec:TD}). \sys~introduces an iteration smoothing algorithm for smoother
\newpage
denoising, hierarchical and credit decoding for enhanced parallel decoding, and a vicinity refresh strategy for KV-cache management to mitigate cache staleness.

Beyond algorithmic improvements, \sys~integrates several system-level optimizations. It supports both tensor parallelism (TP) and expert parallelism (EP) to maximize GPU utilization even at batch size 1. It leverages PyTorch compilation and NVIDIA CUDA Graphs for efficient kernel execution, and introduces a loop unrolling mechanism to eliminate CUDA stream bubbles across diffusion iterations.

\begin{figure}
\centering
\begin{subcaptionblock}{.5\textwidth}
  \centering
  \includegraphics[width=.9\linewidth]{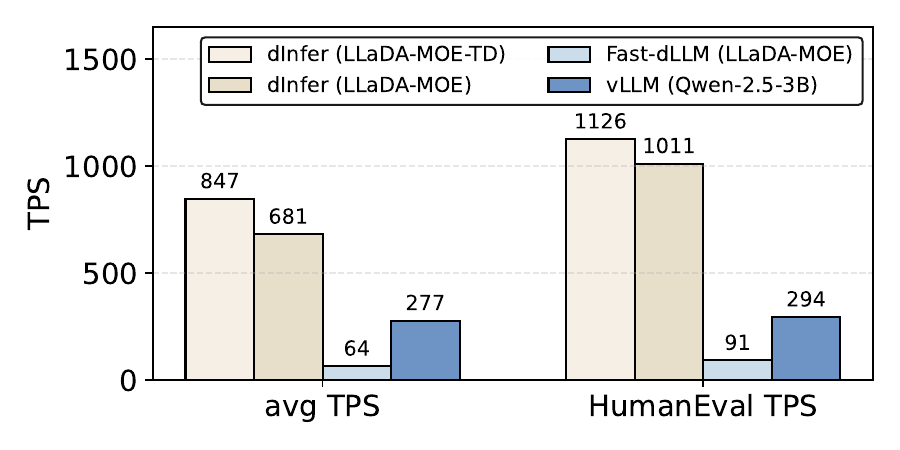}
  \caption*{(a).~TPS}
  \label{fig:sub1}
\end{subcaptionblock}%
\begin{subcaptionblock}{.5\textwidth}
  \centering
  \includegraphics[width=.9\linewidth]{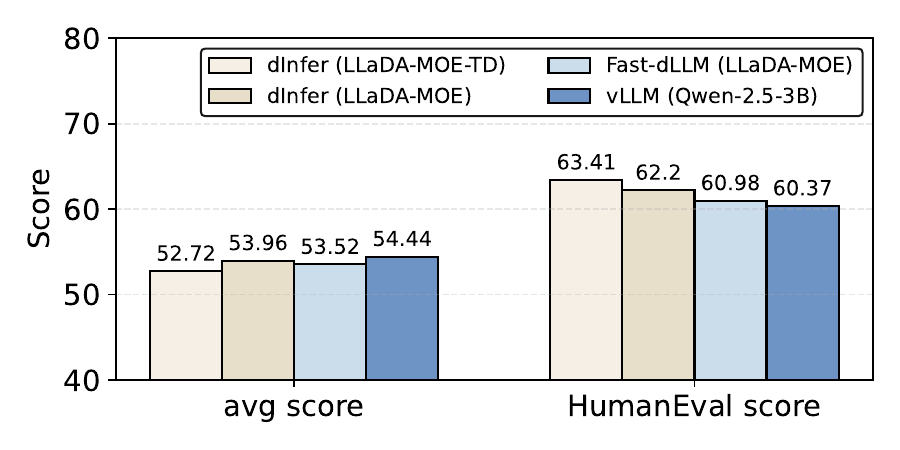}
  \caption*{(b).~Model performance}
  \label{fig:sub2}
\end{subcaptionblock}
\caption{Benchmark results. We compare \sys~on \lladamoe~and \lladamoe-TD with \fastdllm~and vLLM across six benchmarks and show their average inference speed in tokens per second (TPS) on the six benchmarks and the highest inference speed on the HumanEval dataset. When achieving similar model performance on the benchmarks, \sys~is about $10\times$ faster than \fastdllm~on the same model and is about $2-3\times$ faster than vLLM on Qwen-2.5-3B.
}
\label{fig:res}
\end{figure}

We evaluate inference efficiency using tokens per second (TPS) per sequence, shown in Figure \ref{fig:res}. On HumanEval, \sys~achieves over 1,100 TPS at batch size 1, and averages more than 800 TPS across six benchmarks on a single node with $8\times$ H800 GPUs. Compared to \fastdllm~\citep{wu2025fast}, \sys~delivers more than a $10\times$ speedup while maintaining accuracy; on \lladamoe~it provides a $2-3\times$ speedup over QWen2.5-3B on vLLM with comparable quality.

Our contributions are summarized as follows:
\begin{itemize}
    \item We present \sys, the first modularized dLLM inference framework that integrates algorithmic innovations with system-level optimizations to deliver substantial efficiency gains.
    \item We provide the first open-source demonstration that dLLM inference can surpass AR models at batch size 1, establishing a new milestone for inference efficiency. The implementation is available at \url{https://github.com/inclusionAI/dInfer}.
\end{itemize}

\section{Framework Design}

\begin{wrapfigure}{r}{0.5\textwidth}
  \centering
  \includegraphics[width=\linewidth]{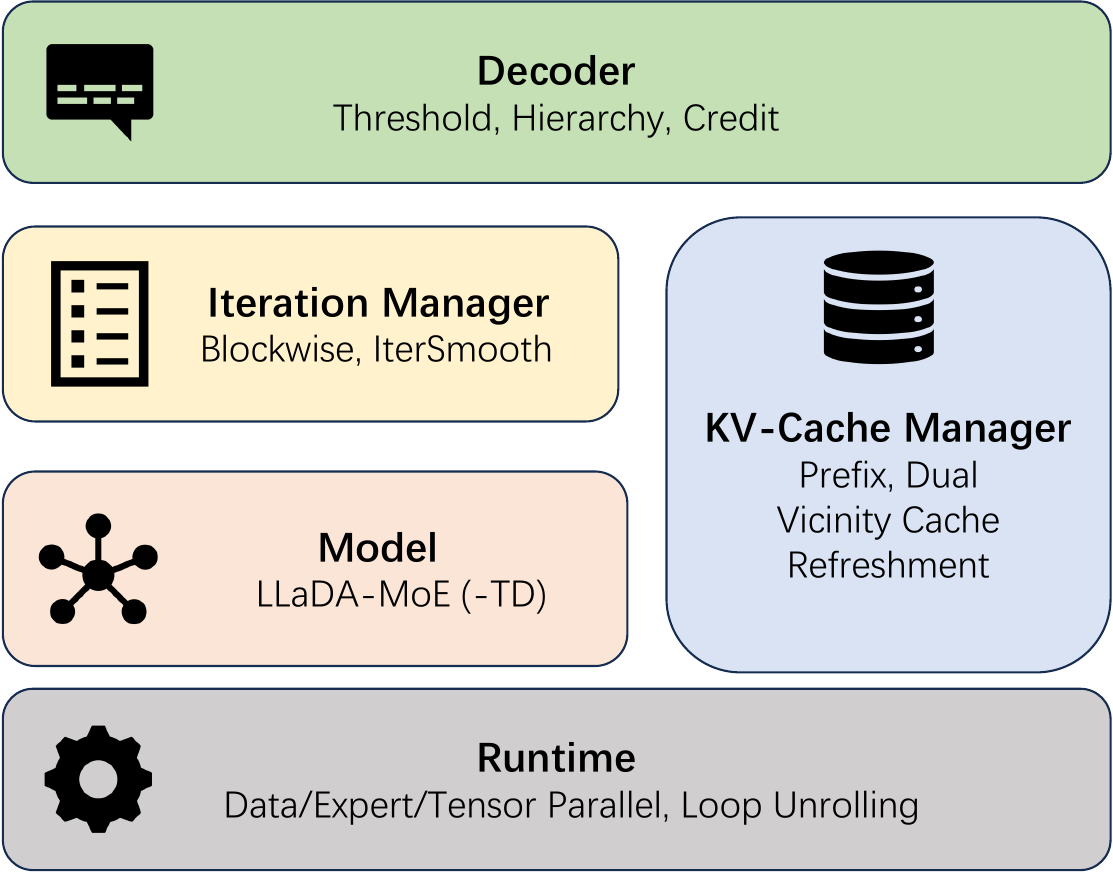}
  \caption{The architecture of the dInfer framework}
  \label{fig:arch}
  \vspace{-30pt}
\end{wrapfigure}


A key advantage of dLLMs lies in their ability to perform parallel decoding within each iteration. To fully exploit this capability, advances are required in model design, diffusion iteration strategies, and decoding algorithms. At the same time, dLLMs introduce unique computational challenges. Unlike AR models--where previously computed keys and values can be cached--dLLMs employ bidirectional attention, meaning that decoding a single token can affect the representations of all tokens in the sequence. This makes straightforward KV-cache reuse infeasible and necessitates specialized cache management for efficient inference.


To address these issues, we design \sys, an inference framework that accelerates dLLMs through four modular components: model, diffusion iteration manager, decoding strategy, and KV-cache management (Figure~\ref{fig:arch}). This modular architecture enables flexible combinations of algorithms across components, allowing users to construct customized inference pipelines that maximize the benefits of parallel decoding while improving computational efficiency.

\begin{algorithm}[H]
\caption{Blockwise dLLM Inference}

\label{alg:blockwise}
\begin{algorithmic}[1]
\Require Input tokens $X \in \mathbb{Z}^{B \times L}$ (undecided positions marked as $mask\_id$);
         block size $S$; model $\mathcal{M}$; decoder $\mathcal{D}$; KV-cache manager $\mathcal{K}$; block iteration manager $\mathcal{I}$
\Ensure  Completed tokens $\widehat{X} \in \mathbb{Z}^{B \times L}$
\State $\mathcal{K}.\textsc{Create}(B, L)$ \Comment{Create KV cache}
\While{$\mathcal{I}.\textsc{HasNext}()$}
\State \textbf{\# 1) Iterator: pick next block (blockwise order)}
\State $[start\!:\!end] \gets \mathcal{I}.\textsc{NextBlock}()$ \Comment{Get next block}
\State $\text{undecided} \gets (X[start:end] = mask\_id)$ \Comment{Boolean mask of undecided positions}
\While{$\text{any}(\text{undecided})$}
  \State \textbf{\# 2) KV update policy}
  \If{$\mathcal{K}.\textsc{ShouldUpdate}(\text{loop\_context}, start\!:\!end)$}
      \State $\mathcal{K}.\textsc{Update}(X, start\!:\!end)$
  \EndIf
  \State \textbf{\# 3) Model forward on this region (with KV Cache)}
  \State $\text{logits} \gets \mathcal{M}.\textsc{Forward}(X, \mathcal{K}, start\!:\!end)$
  \Comment{$\text{logits} \in \mathbb{R}^{B \times L \times V}$}
  \State \textbf{\# 4) Decoder: tokens to commit in this block}
  \State $(X,\, undecided) \gets \mathcal{D}.\textsc{Decode}(\text{logits}, X, \text{undecided}, start\!:\!end)$
\EndWhile
\EndWhile
\State \textbf{return} $X$ \Comment{$\widehat{X}$}
\end{algorithmic}
\end{algorithm}

\subsection{Diffusion Iteration Manager} \label{sec:diff_iter}


The diffusion iteration manager acts as the controller of the iterative denoising process, with three main responsibilities: 1) determining the next region of tokens to decode, 2) interacting with the model to obtain outputs such as logits and hidden states, 3) maintaining historical predictions to provide a richer context for future decoding.

\sys~currently implements two algorithms in this component. The \textbf{blockwise diffusion iteration} algorithm performs decoding in fixed-size spans and serves as a baseline (Algorithm~\ref{alg:blockwise}). The \textbf{iteration smoothing} algorithm improves upon this by retaining token representations from the previous iteration and fusing them with the next iteration’s embeddings. This enables cross-iteration information flow, enriches contextual cues, and empirically boosts token confidence while mitigating the performance degradation typically caused by KV-cache deployment. A detailed description is provided in Appendix \ref{sec:itersmooth}.

\subsection{Decoding Strategy} \label{sec:para_dec}



\sys~supports three strategies for parallel decoding:
\begin{itemize}
\item Threshold decoding (from \fastdllm~\citep{wu2025fast}): commits tokens whose confidence exceeds a preset threshold.
\item Hierarchical decoding (ours): recursively partitions masked spans, ensuring at least one token is decoded per region, thereby reducing local dependencies and improving efficiency.
\item Credit decoding (ours): accumulates historical confidence scores as \textit{credits} and preferentially commits tokens with consistently stable predictions, improving reliability across iterations.
\end{itemize}

These algorithms allow \sys~to achieve higher decoding efficiency without retraining the underlying model. Detailed formulations are included in Appendix \ref{sec:HD} and \ref{sec:CD}.

\subsection{KV-cache management} \label{sec:kvcache}

A central challenge in dLLM inference is KV-cache incompatibility. In AR models, causal attention allows KV states to be computed once and reused; in dLLMs, however, token representations evolve across denoising steps, making static reuse infeasible. Without caching, inference must perform Transformer computations over the entire sequence, creating heavy computational overhead.


Earlier approaches introduced training-free strategies such as blockwise caching and Dual Cache \citep{wu2025fast}, which reuse KV states for decoded tokens or suffixes of masked tokens. However, these methods treat cached states as static, neglecting updates from newly decoded tokens, which often degrades accuracy.


To balance cost and performance, \sys~introduces \textbf{vicinity KV-cache refresh}. This method exploits semantic locality by selectively updating a small window of tokens adjacent to the current decoding block. During denoising, K and V states are recomputed for both masked tokens and their immediate neighbors; once a block is fully decoded, a full cache update ensures global consistency.

\subsection{Model support}

\sys~is designed to be model-agnostic and currently supports state-of-the-art dLLMs such as \lladamoe, LLaDA-1.5, and LLaDA-Instruct, with plans for continued extension. Beyond compatibility, we also explore improving the parallel decoding capability of dLLMs through training. Specifically, we introduce Trajectory Distillation, applied to \lladamoe, yielding the enhanced \lladamoe-TD variant (see Section~\ref{sec:TD}). This method fine-tunes models using effective decoding trajectories identified from their own generation process, significantly boosting parallel decoding efficiency.

\subsection{An example of the orchestration of the algorithms in \sys}

\begin{figure}[h!]
  \centering
\includegraphics[width=0.9\linewidth]{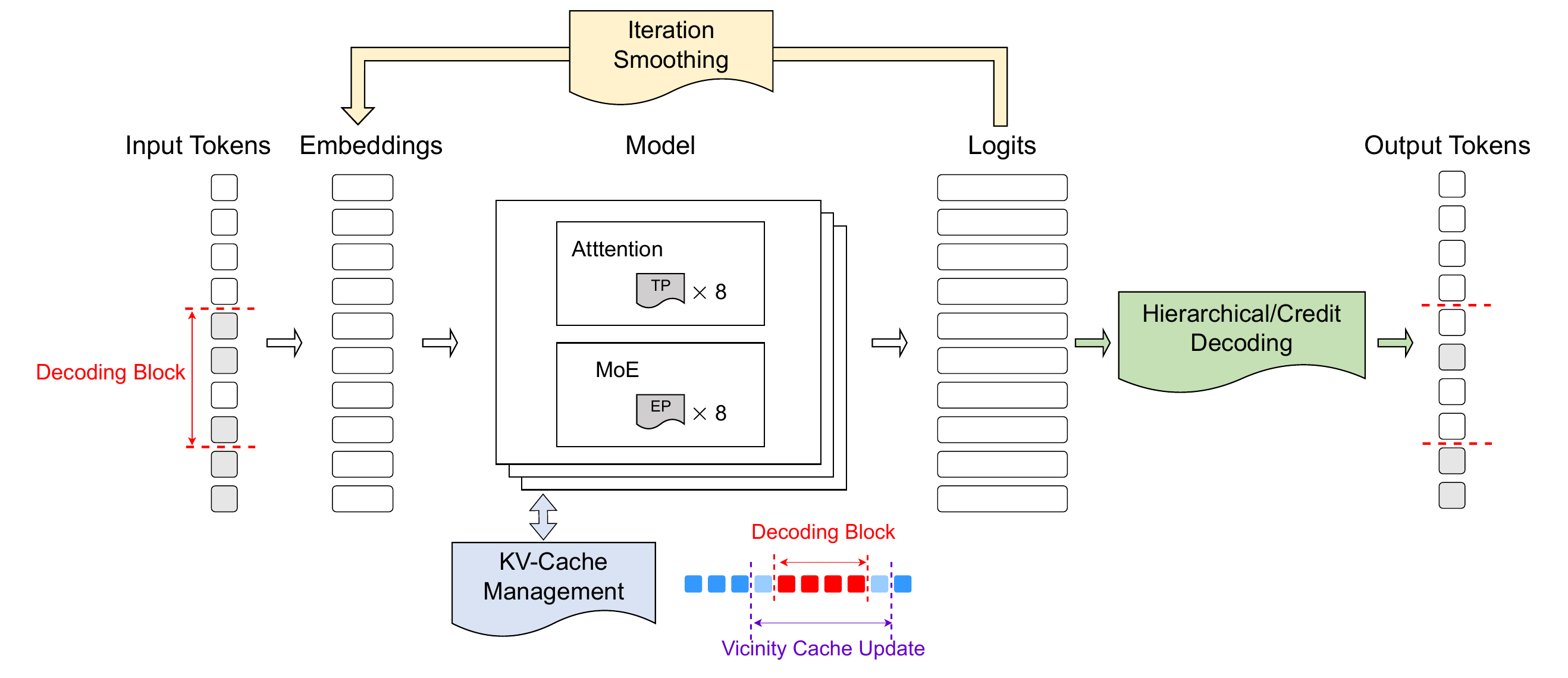}
    \caption{Orchestration of the algorithms in different dInfer components.}
    \label{fig:dInfer_overview}
\end{figure}

Figure \ref{fig:dInfer_overview} shows how the algorithms interact within \sys. 
In each iteration, the framework tries to identify the [MASK] tokens in the active decoding block. The sequence is first embedded (potentially leveraging context from previous iterations), followed by a forward pass through the model using tensor/expert parallelism to produce logits. During the forward pass, the previous KV cache that does not hit the vicinity refresh strategy will remain unchanged and be reused. After obtaining the logits, the hierarchical/credit decoding algorithm will decide which [MASK] tokens to decode and predict their identities. In addition, the iteration smoothing algorithm will retain the logit-weighted embeddings of the sequence and incorporate them as part of the embedding for the next iteration, ensuring continuity across steps. 
Together, these components enable \sys~to achieve both efficiency and stability in dLLM inference.


\section{Implementation Details}

To deliver fast inference speed, \sys~provides system-level optimizations on the components of \sys.


\paragraph{Model computations}


\sys~builds on vLLM’s backend to exploit two complementary forms of parallelism. Tensor parallelism is applied to the linear layers preceding attention modules, distributing dense computations efficiently across multiple GPUs. Expert parallelism is applied to the \lladamoe~model and is effective even at a batch size of 1--unlike in AR models, where expert parallelism typically requires large batch sizes. By combining tensor and expert parallelism, \sys~achieves a more than 100\% improvement in inference efficiency.


To further optimize single-sequence inference, \sys~uses PyTorch's just-in-time (JIT) compiler \texttt{torch.compile} to fuse CUDA kernels and execute them within NVIDIA CUDA Graphs, thereby eliminating PyTorch execution overhead. This compilation technique improves inference efficiency by over 200\% when TP and EP are enabled.

\paragraph{Diffusion iterations}

Diffusion iterations can suffer from CUDA stream bubbles--idle gaps of consecutive kernel launches between diffusion iterations--which waste GPU cycles and reduce throughput. To address this, \sys~applies a loop unrolling strategy that allows Python to launch CUDA kernels continuously without being blocked by stream synchronization. This reduces launch latency, keeps GPU pipelines fully occupied, and boosts iteration efficiency by about 5-10\%.


We also introduce an early termination mechanism for blockwise decoding. Once an end-of-sequence (EOS) token is generated within a block, subsequent decoding steps on the remaining blocks become redundant. \sys~therefore halts the diffusion loop and fills all remaining blocks with EOS, avoiding unnecessary computation. This optimization improves inference efficiency by 15-40\%.

\paragraph{Parallel decoding}
To make loop unrolling effective in diffusion iterations, decoding algorithms in \sys~are implemented without control-flow operations, and data transfer from PyTorch tensors to Python code is eliminated. This design ensures compatibility with system-level optimizations to achieve high decoding throughput.

\section{Evaluations} \label{sec:eval}

\subsection{Datasets and Configurations}

\textbf{Datasets.} We select six datasets from diverse domains with sufficient response lengths: CRUX-O~\citep{gu2024cruxeval}, LiveCodeBenchv6~\citep{jain2024livecodebench}(denoted as LCB V6), MBPP~\citep{austin2021program}, and HumanEval~\citep{chen2021evaluating} for code generation; GSM8K~\citep{cobbe2021training} for mathematical reasoning; and IFEval~\citep{zhou2023instruction} for instruction-following agent tasks.

\textbf{Evaluation Metric.} To evaluate the efficiency of the \sys~framework, we use \textbf{tokens per forward (TPF) per sequence} to measure the parallel decoding capability within a single diffusion iteration, and \textbf{tokens per second (TPS) per sequence} to assess overall inference efficiency.
To be more specific, 
TPF can be formally described as
$TPF=\frac{1}{N}\sum_{i \in D}\frac{T_i}{F_i}$, where $T_i$ and $F_i$ are the number of tokens generated before the first \texttt{EOS} and the number of diffusion iterations that run on a sequence $i$ to generate tokens, and $D$ is the dataset. TPS can be described as $TPS=\frac{1}{N}\sum_{i \in D}\frac{T_i}{t_i}$ where $t_i$ is the time cost to generate tokens for sequence $i$.


\textbf{Configurations.}
We compare \sys~with \fastdllm~\citep{wu2025fast} to demonstrate the effectiveness of both system optimizations and algorithmic innovations in \sys.
%
Without KV cache, \fastdllm~employs parallel decoding with a threshold of 0.9, a setting used in their paper, while \sys~further incorporates \textit{credit decoding} and \textit{iteration smoothing} to validate the effectiveness of the proposed decoder. 
When KV cache is enabled, \fastdllm~adopts Dual Cache, whereas \sys~integrates the \textit{vicinity KV-Cache refresh} method with \textit{iteration smoothing} and threshold decoding. 
In addition, we evaluate the effectiveness of \lladamoe-TD in \sys. We report its results under the dInfer optimal setting, which integrates dual-cache, \textit{hierarchical decoding}, \textit{vicinity KV-Cache refresh}, and \textit{iteration smoothing}. Please see more details about the configurations of the experiments in Appendix \ref{sec:config}.


    

All experiments are conducted on a server equipped with $8\times$ NVIDIA H800 GPUs, with PyTorch 2.9.0.dev20250831 and vLLM 0.10.1. 
We use a batch size of 1, a generation length of 1024 and a block size of 64 for all experiments.


\begin{table}[ht]
\footnotesize
\caption{Evaluations of different framework and configurations in terms of performance, TPF, and TPS on \lladamoe. We can observe that dInfer achieves a $2-3 \times$ improvement over vLLM (680.71 vs. 277.45). Furthermore, we can see that dInfer provides more than a tenfold enhancement over \fastdllm~(680.71 vs. 63.61) while achieving similar results (53.96 vs. 53.52).
}
\label{tab-perf}

\begin{tabular}{c|c|c|c|c|c|c|c|c|c}
\toprule
Config. & Frame. & \multicolumn{1}{c|}{Metric} & Avg. & CRUX-O & GSM8K & HumanEval & IFEval & MBPP & LCB V6 \\ \midrule
\lladamoe & - & Perf & 54.83 & 42.38 & 82.41 & 61.59 & 59.33 & 70.02 & 13.27 \\ \midrule
\multirow{3}{*}{QWen2.5-3B} & \multirow{3}{*}{vLLM} & Perf & 54.44 & 46.75 & 86.28 & 60.37 & 58.2 & 65.81 & 9.2 \\
 &  & TPF & 1 & 1 & 1 & 1 & 1 & 1 & 1 \\
 &  & TPS & 277.45 & 289.53 & 294.15 & 294.05 & 296.7 & 290.15 & 200.12 \\ \midrule
\multirow{6}{*}{\begin{tabular}[c]{@{}c@{}}Without \\ \\ KV Cache\end{tabular}} & \multirow{3}{*}{\fastdllm} & Perf & 53.52 & 43.75 & 82.79 & 60.98 & 54.53 & 66.5 & 12.56 \\
 &  & TPF & 2.82 & 2.9 & 2.28 & 3.87 & 2.42 & 3.01 & 2.46 \\
 &  & TPS & 63.61 & 59.79 & 56.19 & 90.8 & 60.25 & 70.2 & 44.4 \\ \cmidrule{2-10}
 & \multirow{3}{*}{\sys} & Perf & \textbf{54.33} & 42.38 & 82.26 & 63.41 & 57.49 & 67.21 & 13.22 \\
 &  & TPF & \textbf{4.29} & 4.26 & 3.76 & 6.17 & 2.79 & 4.82 & 3.92 \\
 &  & TPS & \textbf{407.36} & 379.62 & 379.63 & 606.85 & 285.49 & 475.23 & 317.36 \\ \midrule
\multirow{6}{*}{\begin{tabular}[c]{@{}c@{}}With\\ \\ KV Cache\end{tabular}} & \multirow{3}{*}{\fastdllm} & Perf & 52.15 & 40.75 & 79.9 & 60.37 & 53.97 & 65.11 & 12.78 \\
 &  & TPF & 2.46 & 2.68 & 2.09 & 3.24 & 2.02 & 2.55 & 2.19 \\
 &  & TPS & 110.98 & 120.57 & 97.5 & 143.9 & 95.23 & 112.9 & 95.8 \\ \cmidrule{2-10}
 & \multirow{3}{*}{\sys} & Perf & \textbf{53.96} & 41.38 & 80.97 & 62.2 & 58.78 & 67.45 & 13 \\
 &  & TPF & \textbf{3.87} & 4.02 & 3.42 & 5.52 & 2.32 & 4.54 & 3.38 \\
 &  & TPS & \textbf{680.71} & 765.3 & 682.9 & 1,011.12 & 444.51 & 757.55 & 422.88 \\ \bottomrule
\end{tabular}
\end{table}

\begin{table}[ht]
\footnotesize
\caption{Evaluations of different framework and configurations in terms of performance, TPF, and TPS on \lladamoe-TD. With the introduction of Trajectory Distillation, the TPS for various benchmarks has significantly improved. The average TPS exceeds that of vLLM by more than threefold.}
\label{tab-perf-td}
\begin{tabular}{c|c|c|c|c|c|c|c|c|c}
\toprule
Config. & Frame. & Metric & Avg. & CRUX-O & GSM8K & HumanEval & IFEval & MBPP & LCB V6 \\ \midrule
\lladamoe & - & Perf & 54.83 & 42.38 & 82.41 & 61.59 & 59.33 & 70.02 & 13.27 \\ \midrule
\multirow{3}{*}{QWen2.5-3B} & \multirow{3}{*}{vLLM} & Perf & 54.44 & 46.75 & 86.28 & 60.37 & 58.2 & 65.81 & 9.2 \\
 &  & TPF & 1 & 1 & 1 & 1 & 1 & 1 & 1 \\
 &  & TPS & 277.45 & 289.53 & 294.15 & 294.05 & 296.7 & 290.15 & 200.12 \\  \midrule
\multirow{3}{*}{\begin{tabular}[c]{@{}c@{}}With\\ KV Cache\end{tabular}} & \multirow{3}{*}{\sys} & Perf & \textbf{52.72} & 40.12 & 79.15 & 63.41 & 56.19 & 65.11 & 12.33 \\
 &  & TPF & \textbf{5.67} & 6.06 & 6.12 & 7.10 & 2.98 & 6.61 & 5.18 \\
 &  & TPS & \textbf{847.22} & 976.66 & 1,011.22 & 1,125.67 & 496.92 & 906.98 & 562.87 \\ \bottomrule
\end{tabular}
\end{table}

\subsection{Performance}
As shown in Table~\ref{tab-perf}, it can be clearly observed that \lladamoe~has comparable or higher performances than Qwen2.5-3B over six different datasets.
Under the "Without KV Cache" setting, \sys~achieves an average accuracy of 54.33, which is higher than \fastdllm~and is comparable to the performance of QWen2.5-3B in vLLM and \lladamoe~reported in its paper \citep{zhu2025lladamoe}, while achieving $6.5\times$ speedup over \fastdllm.
When KV cache is enabled, \sys~achieves higher accuracy than \fastdllm~(53.96 vs. 52.15) and delivers $6\times$ speedup over \fastdllm. 
When achieving similar model performance, \sys~achieves over $10 \times$ speedup (a TPS of 680.71) over \fastdllm~(a TPS of 63.61) across the six benchmarks. \sys~is also $2.5 \times$ faster than Qwen-2.5 3B (a TPS of 277.45) in vLLMs.

 As shown in Table~\ref{tab-perf-td}, the \lladamoe~model trained by the trajectory distillation technique substantially improves inference efficiency on the six benchmarks. Coupled with the full set of \sys~algorithmic optimizations, the distilled model achieves an average TPS of 847.22, significantly higher than the 680.71 TPS of the non-distilled baseline, which is more than \textbf{3x} speedup over Qwen2.5-3B in vLLM.

\section{Conclusion}
In this work, we present \sys, an efficient and extensible inference framework for dLLMs. By decomposing inference into modular components and incorporating optimizations such as hierarchical decoding, credit decoding, iteration smoothing, and vicinity KV-cache refresh, \sys~effectively addresses the key bottlenecks of high computation cost and limited parallel decoding efficiency of dLLMs. Extensive experiments on \lladamoe~demonstrate that \sys~achieves state-of-the-art throughput--exceeding 1,100 TPS on $8\times$H800 GPUs--while maintaining output quality. We believe that \sys~provides both a practical toolkit and a standardized platform to accelerate research and development in the rapidly growing field of dLLMs.

\bibliographystyle{antgroup}
\bibliography{ref/Top,ref/reference}

\clearpage
\appendix

\section{Diffusion iteration}

\subsection{IterSmooth: Iteration Smoothing} \label{sec:itersmooth}

In conventional dLLM decoding, only a small subset of positions is updated at each step by selecting argmax tokens, while the logits for all other positions are discarded. IterSmooth reuses this otherwise wasted information: for positions that remain masked, it converts the logits distribution into an expected embedding and injects it into the corresponding mask-token embedding. This allows uncertain positions to be enriched with dense, distribution-level signals. In our setting, since the output projection and input embedding matrices are not weight-tied, the expected embeddings are computed using the input embedding matrix together with the token probability distribution.

We operate only on masked positions to avoid shifting the training distribution elsewhere. Let $z_t[i]$ be the logits at step $t$ and position $i$, $W_{\mathrm{emb}}$ the input embedding matrix, and $e_{\mathrm{mask}}$ the standard mask embedding. Without temperature scaling, we use:
\[
p_t[i] = \mathrm{softmax}(z_t[i]), \quad
\Delta e_t[i] = p_t[i]\, W_{\mathrm{emb}}, \quad
e_{t+1}[i] = e_{\mathrm{mask}} + \alpha_t \,\Delta e_t[i], 
\quad
\alpha_t = min(\alpha_{init} + \alpha_{growth}t, \alpha_{preset})
\]

The mixing weight $\alpha_t$ increases from a small initial value (e.g., 0.1) over decoding steps toward a preset maximum (e.g., 0.2–0.4), ensuring conservative behavior early.
In addition, we adopt a decode-threshold schedule that decays from 1.0 toward a preset target across steps, decoding only high-confidence positions early while progressively relaxing the criterion, which stabilizes inputs early and increases the contribution of distribution-level guidance later. 
Our method does not introduce new parameters or retraining, and $W_{\mathrm{emb}}$ is reused directly.

The approach increases per-step information by leveraging the full distribution instead of only argmax tokens. Our empirical study shows that this method can increase the average number of tokens decoded in a diffusion iteration by $30-40\%$, and improves the final quality of generated texts. 


\section{Decoding strategy}

\subsection{Hierarchical decoding}
\label{sec:HD}
While dLLMs theoretically allow parallel decoding by predicting multiple tokens at once, naive implementations often suffer from quality degradation. This stems from the violation of the conditional independence assumption among simultaneously generated tokens, which frequently leads to semantic inconsistencies.

To overcome this limitation, we propose Hierarchical Decoding, a training-free strategy inspired by the divide-and-conquer paradigm. The method recursively partitions masked spans into smaller sub-regions and decodes tokens based on their confidence, attempting to resolve at least one token in each region during every forward pass whenever confidence permits. The key insight is that the spatial distribution of masked tokens has a critical impact on prediction stability.

This approach provides two notable advantages. First, by promoting non-contiguous decoding, it increases the spacing between masked tokens, thereby reducing local dependencies and improving semantic consistency. Second, when decoding positions are preferentially selected near the center of each span, the undecoded regions shrink recursively, enabling the process to approach $O(log n)$ complexity in the ideal case. Together, these properties allow Hierarchical Decoding to generate more tokens per forward pass than vanilla decoding without fine-tuning the base model.

\subsection{Credit decoding}
\label{sec:CD}
In standard dLLM inference, text is generated through repeated \textit{predict–sample–re-mask} cycles across multiple denoising steps. Existing parallel decoding strategies typically commit tokens based solely on their \emph{current} confidence at each step. In practice, however, many tokens that are ultimately correct stabilize early in the generation process but remain below the confidence threshold. These tokens are repeatedly re-masked and re-evaluated, leading to unnecessary computation. We present \emph{CreditDecoding}, a training-free acceleration algorithm for dLLMs that reduces redundant computation and accelerates convergence in parallel decoding. 

During decoding, we maintain a \emph{credit} $C_t^{i,v}$ for each position $i$ and token $v \in \mathcal{V}$. This credit which quantifies how consistently a token has been favored along the generation process, serving as a temporal prior for its likelihood of being correct. Given the input $x_t$, let $p_\theta^i(v \mid x_t) = \mathrm{Softmax}(f_\theta(x_t)^i_v)$ denote the model’s current predictive distribution, where $f_\theta(x_t)$ are the logits. Let $v^* = \arg\max_v p_\theta^i(v \mid x_t)$ be the top candidate token at position $i$. The credit is updated as follows:
\begin{equation}
\label{eq:credit-update-top1}
C_t^{i,v}=
\begin{cases}
\beta\, C_{t-1}^{i,v} + \big(p_\theta^i(v\mid x_t)\big)^{\gamma} & v=v^*,\\[3pt]
\beta\, C_{t-1}^{i,v} & \text{otherwise},
\end{cases}
\qquad \beta\in(0,1),\;\gamma\in(0,1).
\end{equation}
Here $\beta$ discounts the earlier confidence and prevents errors from accumulating and influencing future prediction. The concave transformation $(\cdot)^\gamma$ (with $\gamma < 1$) provides relatively larger boosts to tokens with low or moderate confidence, helping correct but underconfident predictions stabilize earlier. 

Before making a token commitment decision, the accumulated credit is fused with the model’s logits as a prior in the log domain:
\begin{equation}
\label{eq:logits-fuse}
\tilde f_\theta(x_t)^i_v = f_\theta(x_t)^i_v + \alpha\,\log\!\big(1 + C_t^{i,v}\big),\qquad \alpha > 0,
\end{equation}
which yields an enhanced distribution $\tilde p_\theta^i(v \mid x_t) = \mathrm{Softmax}\!\big(\tilde f_\theta(x_t)^i_v\big)$. Intuitively, tokens that have been consistently predicted across steps receive a confidence boost, making them more likely to be committed earlier. In contrast, tokens with fluctuating or transiently high confidence are suppressed. This mechanism enhances decoding stability, particularly in long-sequence and reasoning tasks.


Importantly, CreditDecoding does not change the underlying sampling or decoding policy, but instead simply replaces the original distribution $p_\theta$ with the enhanced $\tilde p_\theta$. This design ensures the compatibility with standard inference optimizations such as  threshold decoding, top-$k$ sampling, KV-cache, and compiler-level optimizations, allowing efficiency gains to accumulate when combined.

To balance efficiency and robustness under varying stability conditions, we default to maintaining and updating credits only within the current decoding block. This limits the influence of uncertain future context, reduces interference from under-informed positions, and improves scalability across different model sizes and context lengths—especially in long-sequence generation scenarios.


\section{Post-training to enhance models' parallel decoding}

\subsection{Inference Acceleration via Trajectory Compression} \label{sec:TD}

While dLLMs show promise for non-sequential generation, their practical application is often hindered by high inference latency stemming from the iterative, multi-step sampling process. Inspired by the approach in Seed Diffusion \citep{song2025seed}, which demonstrates the value of training on high-quality generation paths, we propose a novel second-stage fine-tuning method, termed \textbf{Trajectory Compression}, to explicitly reduce the number of required sampling steps. The core idea is to train the model to "jump" between non-consecutive states within an optimal generation trajectory, thereby decoding multiple tokens in a single forward pass. We refer to this resulting model as \lladamoe-TD.

Our method consists of two main stages: high-quality trajectory distillation and compressed transition learning.

\paragraph{High-Quality Trajectory Distillation}
First, we generate a dataset of "golden" trajectories. We use a pre-trained dLLM to sample a large corpus of generation paths, $\mathcal{T}$, on a domain-specific dataset (e.g., 200,000 math problems). A trajectory $\tau = (s_N, s_{N-1}, \dots, s_0)$ represents the sequence of states from the initial fully masked sequence $s_N$ to the final generated output $s_0$.

Each trajectory's final output $s_0$ is evaluated by an external verifier, $V(\cdot)$. For mathematical tasks, we employ a \textit{math\_verify} function to ascertain the correctness of the solution. We then filter the corpus to retain only the trajectories that result in a correct output, forming a high-quality dataset $\mathcal{T}_{\text{gold}}$:
$$
\mathcal{T}_{\text{gold}} = \{ \tau \in \mathcal{T} \,|\, V(s_0^{\tau}) = \text{True} \}
$$
This process ensures that the subsequent fine-tuning stage learns from effective and valid reasoning paths.

\paragraph{Compressed Transition Learning}
In the second stage, we fine-tune the dLLM on a new objective. Instead of learning the standard single-step transition $p_\theta(s_{t-1} | s_t)$, we train the model to predict a multi-step transition from an early state $s_i$ to a later state $s_j$, where $i > j$.

For each trajectory $\tau \in \mathcal{T}_{\text{gold}}$, we construct a training instance by randomly sampling two timestamps, $i$ and $j$, where $N \ge i > j \ge 0$. The pair $(s_i, s_j)$ serves as the input and target, respectively. Since tokens, once revealed, are fixed in subsequent steps, the model's task is to predict the tokens that are `[MASK]` in $s_i$ but are revealed in $s_j$. Let $M_t$ be the set of indices of `[MASK]` tokens in state $s_t$. The model learns to predict the tokens at indices $\Delta_{i \to j} = M_i \setminus M_j$.

The fine-tuning objective is to minimize the negative log-likelihood of this compressed transition. The loss function is defined as:
$$
\mathcal{L}_{\text{compress}}(\theta) = - \mathbb{E}_{\tau \in \mathcal{T}_{\text{gold}}, \, i,j \sim U(\tau)} \left[ \sum_{k \in \Delta_{i \to j}} \log p_\theta(x_k = s_j[k] \,|\, s_i) \right]
$$
where $s_j[k]$ is the ground-truth token at position $k$ in the target state $s_j$. To handle variable-length sequences, both $s_i$ and $s_j$ are padded to the model's maximum context length.

This fine-tuning process endows the model with the ability to execute large  \textit{jumps} during inference, significantly accelerating generation. We measure this improvement using tokens per forward (TPF). Our experiments show that this method yields a 99.8\% increase in TPF for mathematical reasoning and an average TPF improvement of 45.3\% across other domains, including code generation, confirming Trajectory Compression as an effective technique for reducing dLLM inference latency.

\section{Detailed Configuration of Experiments} \label{sec:config}
To achieve an optimal balance between efficiency and generation quality, different experimental settings adopt distinct combinations of decoding and optimization methods, as summarized in Table~\ref{tab:exp_settings}. Each configuration is tuned for its best trade-off between model performance and inference efficiency given the model’s characteristics and cache usage. 
We thus employ tailored algorithms for each setting, based on the ablation results provided in Table~\ref{tab:ab-wocache}, Table~\ref{tab:ab-cache}, and Table~\ref{tab:ab-cachetd}, respectively.

The hyperparameter settings for the corresponding methods are as follows:
the threshold decoding uses a confidence threshold of 0.8;
the hierarchical decoding adopts a decoding threshold of 0.92 and a lower boundary threshold of 0.62;
the iteration smoothing employs a continuation weight (cont\_weight) of 0.3;
and the vicinity KV-Cache refreshment strategy uses a prefix look and after look of 16, with warmup\_times = 4.

\begin{table}[h!]
\centering
\caption{Experimental settings and enabled methods. A checkmark (\checkmark) indicates the method is applied.}
\label{tab:exp_settings}
\renewcommand{\arraystretch}{1.15}
\setlength{\tabcolsep}{5pt}
\begin{tabular}{lccccc}
\toprule
\textbf{dInfer Setting} & \textbf{Threshold} & \textbf{Hier.} & \textbf{Credit} & \textbf{IterSmooth.} & \textbf{Vicinity KV-Ref.} \\
\midrule
\lladamoe~w/o KV-Cache       &  $\times$ & $\times$ &  \checkmark& \checkmark & $\times$ \\
\lladamoe~with KV-Cache      & \checkmark & $\times$ & $\times$ & \checkmark & \checkmark \\
\lladamoe-TD with KV-Cache   &  $\times$ & \checkmark & $\times$ & \checkmark & \checkmark \\
\bottomrule
\end{tabular}
\end{table}

\begin{table}[htbp]
\centering
\caption{Ablation study of decoding algorithm on \lladamoe~w/o KV-Cache setting.}
\label{tab:ab-wocache}
\begin{tabular}{c|c|c|c|c|c|c|c|c}
\toprule
Config. & Metric & Avg. & CRUX-O & GSM8K & HumanEval & IFEval & MBPP & LCB V6 \\ \toprule
\multirow{2}{*}{\begin{tabular}[c]{@{}c@{}}Threshold\\ \end{tabular}} & Perf & 54.01 & 42.62 & 82.41 & 60.98 & 55.64 & 67.21 & 15.2 \\ 
 & TPF & 3.67 & 3.03 & 3.11 & 5.40 & 2.64 & 4.44 & 3.37 \\  
 \midrule
\multirow{2}{*}{\begin{tabular}[c]{@{}c@{}}Hierarchy\\ \end{tabular}} & Perf & 53.68 & 39.75 & 80.97 & 64.02 & 57.67 & 65.81 & 13.88 \\  
 & TPF & 3.89 & 3.28 & 3.55 & 5.80 & 2.34 & 4.75 & 3.64 \\  
 \midrule
\multirow{2}{*}{\begin{tabular}[c]{@{}c@{}}Credit\\ \end{tabular}} & Perf & 54.33 & 42.38 & 82.26 & 63.41 & 57.49 & 67.21 & 13.22 \\  
 & TPF & 4.29 & 4.26 & 3.76 & 6.17 & 2.79 & 4.82 & 3.92 \\  
 \midrule
\end{tabular}
\end{table}

\begin{table}[htbp]
\centering
\caption{Ablation study of decoding algorithm on \lladamoe~with KV-Cache setting.}
\label{tab:ab-cache}
\begin{tabular}{c|c|c|c|c|c|c|c|c}
\toprule
Config. & Metric & \multicolumn{1}{c|}{Avg.} & \multicolumn{1}{c|}{CRUX-O} & \multicolumn{1}{c|}{GSM8K} & \multicolumn{1}{c|}{HumanEval} & \multicolumn{1}{c|}{IFEval} & \multicolumn{1}{c|}{MBPP} & \multicolumn{1}{c}{LCB V6} \\  \midrule
\multirow{2}{*}{\begin{tabular}[c]{@{}c@{}}Threshold\\ \end{tabular}} & Perf & 53.96 & 41.38 & 80.97 & 62.2 & 58.78 & 67.45 & 13 \\  
 & TPF & 3.87 & 4.02 & 3.42 & 5.52 & 2.32 & 4.54 & 3.38 \\  
 \midrule
\multirow{2}{*}{\begin{tabular}[c]{@{}c@{}}Hierarchy\\ \end{tabular}} & Perf & 53.90 & 48.82 & 79.3 & 64.02 & 53.97 & 66.28 & 11.01 \\  
 & TPF & 3.15 & 3.20 & 2.91 & 4.39 & 1.92 & 3.68 & 2.80 \\  
 \midrule
\multirow{2}{*}{\begin{tabular}[c]{@{}c@{}}Credit\\ \end{tabular}} & Perf & 51.56	&37.88	&80.14	&61.59	&53.97	&63.93	&11.87 \\  
 & TPF & 3.4	&3.01	&3.19	&5.01	&2.08	&3.99	&3.12\\  
 \bottomrule
\end{tabular}
\end{table}
\clearpage
\begin{table}[H]
\centering
\caption{Ablation study of decoding algorithm on \lladamoe-TD with KV-Cache setting.}
\label{tab:ab-cachetd}
\begin{tabular}{c|c|c|c|c|c|c|c|c}
\toprule
Config. & Metric & \multicolumn{1}{c|}{Avg.} & \multicolumn{1}{c|}{CRUX-O} & \multicolumn{1}{c|}{GSM8K} & \multicolumn{1}{c|}{HumanEval} & \multicolumn{1}{c|}{IFEval} & \multicolumn{1}{c|}{MBPP} & \multicolumn{1}{c}{LCB V6} \\  \midrule
\multirow{2}{*}{\begin{tabular}[c]{@{}c@{}}Threshold\\ \end{tabular}} & Perf & 49.56 & 35.25 & 76.88 & 57.93 & 56.93 & 60.89 & 9.47 \\   
 & TPF & 5.83 & 6.26 & 6.33 & 7.27 & 3.14 & 6.73 & 5.23 \\   
 \midrule
\multirow{2}{*}{\begin{tabular}[c]{@{}c@{}}Hierarchy\\ \end{tabular}} & Perf & 52.72 & 40.12 & 79.15 & 63.41 & 56.19 & 65.11 & 12.33 \\   
 & TPF & 5.67 & 6.06 & 6.12 & 7.09 & 2.98 & 6.60 & 5.18 \\   
 \midrule
\multirow{2}{*}{\begin{tabular}[c]{@{}c@{}}Credit\\ \end{tabular}} & Perf &48.89	&34.5	&77.18	&57.32	&50.83	&62.53	&10.96 \\   
 & TPF &5.17	&4.75	&5.9	&6.77	&2.77	&6.07	&4.73 \\   
 \bottomrule
\end{tabular}
\end{table}

\end{document}